\documentclass[runningheads]{llncs}

\usepackage[T1]{fontenc}
\usepackage{graphicx}
\usepackage{graphicx}
\usepackage{amsmath, amssymb}
\usepackage{booktabs}
\usepackage{multirow}
\usepackage{caption}
\usepackage{subcaption}
\usepackage{float}
\usepackage{xcolor}
\usepackage{hyperref}
\usepackage{url}
\usepackage{algorithm}
\usepackage{algpseudocode}
\usepackage{listings}
\usepackage{longtable}
\usepackage{bm}
\usepackage{array}
\usepackage{tikz}
\usetikzlibrary{positioning}

\usepackage{devanagari} 

 \expandafter\def\expandafter\normalsize\expandafter{%
    \normalsize
    \setlength\abovedisplayskip{3pt}
    \setlength\belowdisplayskip{2pt}
    \setlength\abovedisplayshortskip{1pt}
    \setlength\belowdisplayshortskip{1pt}
}

\newcolumntype{L}[1]{>{\raggedright\let\newline\\\arraybackslash\hspace{0pt}}m{#1}}
\newcolumntype{C}[1]{>{\centering\let\newline\\\arraybackslash\hspace{0pt}}m{#1}}
\newcolumntype{R}[1]{>{\raggedleft\let\newline\\\arraybackslash\hspace{0pt}}m{#1}}

\begin{document}
\title{Historic Scripts to Modern Vision:  A Novel Dataset and A VLM Framework for Transliteration of Modi Script to Devanagari}
\titlerunning{Historic Scripts to Modern Vision}

\author{Harshal Kausadikar\inst{1}\orcidID{0009-0002-0360-2237} \and
Tanvi Kale\inst{1}\orcidID{0009-0006-4462-8010} \and
Onkar Susladkar\inst{2}\orcidID{0000-0003-4511-1858} \and
Sparsh Mittal \inst{3}\orcidID{0000-0002-2908-993X}
}
\institute{Independent Researcher \email{\{harshalkausadikar,ktanvarada\}@gmail.com} \and Yellow.AI  \email{onkarsus13@gmail.com} \and IIT Roorkee \email{sparsh.mittal@ece.iitr.ac.in}\\Harshal and Tanvi are co-first authors.}

%
%
\maketitle              
\thispagestyle{plain}
\footnotetext[1]{\hfill \textit{The MoDeTrans dataset is available at \url{https://huggingface.co/datasets/historyHulk/MoDeTrans} and SynthMoDe dataset is available at \url{https://huggingface.co/datasets/historyHulk/SynthMoDe}.
We are also open-sourcing both the model and codebase. To streamline accessibility and integration, we are collaborating with Hugging Face’s Transformer libraries, ensuring that researchers and practitioners can easily access the model and its training utilities.
}}

\begin{abstract}
In medieval India, the Marathi language was written using the Modi script. The texts written in Modi script include extensive knowledge about medieval sciences, medicines, land records and authentic evidence about Indian history. Around 40 million documents are in poor condition and have not yet been transliterated. Furthermore, only a few experts in this domain can transliterate this script into English or Devanagari. Most of the past research predominantly focuses on individual character recognition. A system that can transliterate Modi script documents to Devanagari script is needed. We propose  the MoDeTrans dataset, comprising 2,043 images of Modi script documents accompanied by their corresponding textual transliterations in Devanagari. We further introduce MoScNet (\textbf{Mo}di \textbf{Sc}ript \textbf{Net}work), a novel Vision-Language Model (VLM) framework for transliterating Modi script images into Devanagari text. MoScNet leverages Knowledge Distillation, where a student model learns from a teacher model to enhance transliteration performance. 
The final student model of MoScNet has better performance than the teacher model while having 163$\times$ lower parameters.  Our work is the first to perform direct transliteration from the handwritten Modi script to the Devanagari script. MoScNet also shows competitive results on the optical character recognition (OCR) task. 

\end{abstract}

\section{Introduction}
\textbf{A brief history of Modi script:}  Modi script was the official script used during the reign of Marathas for writing the Marathi language.   
It was used in Maharashtra and throughout many other parts of India. 
According to \cite{pandey2011modi}, Modi script can be classified into six types -  Adyakalin Modi (of 12th century), Yadavakalin Modi (of 13th century), Bahmanikalin Modi (of 14th – 16th century), Shivakalin Modi (of 17th century), Peshwekalin Modi (of 18th - 19th century) and Anglakalin Modi (from 1818 till 1952).
Modi script was in use till India's pre-independence era, but later, the script was superseded by the Devanagari script.

\textbf{Importance of Modi script:} In medieval India, the Modi script was widely used for scriptures, official documents, and administrative records, covering topics like land, property, yoga, spirituality, and \textit{Ayurvedic} medicine. These fragile records hold valuable insights into medieval history and science.  
Developing a framework to convert Modi script into recognized scripts would aid transcription, digitization, and research, benefiting historians and people at large. With 83 million Marathi speakers in India (2011 census \cite{census2011}), reviving the Modi script could enhance cultural literacy and regional identity.

\begin{figure}[ht] 
    \centering
        \includegraphics[width=\textwidth]{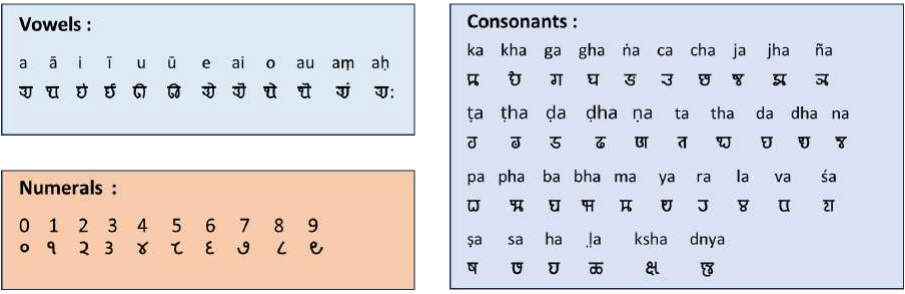}
        \caption{Modi script characters - 12 vowels, 10 numerals, and 36 consonants.}
    \label{fig:modi_characters}
\end{figure}

\textbf{Challenges in transliterating Modi script:}  As shown in  Figure~\ref{fig:modi_characters}, Modi script consists of 48 letters, including 36 consonants and 12 vowels. Certain characters have a striking resemblance in appearance. Modi script has some peculiar characteristics which make its transliteration challenging:
\newline (1) First, the \textit{Shirorekha}, a horizontal line equal to the width of the paper, is  drawn. Then, a letter is written in a way that it starts and ends at the shirorekha. 
\newline (2) Its cursive design was intended for faster writing, resulting in no spaces between words and no clear demarcation of word boundaries.
\newline (3) Ligatures and angular nature. \newline (4) Very few punctuation marks are observed in the historical texts of Modi script.

\textbf{Contributions:} Our contributions are as follows:

1. Introducing MoDeTrans Dataset: Our first significant contribution is the \textbf{MoDeTrans Dataset}, which comprises 2043 Modi script images and their corresponding Devanagari text transliterations, thereby establishing a valuable benchmark for script transliteration research. Most existing works on Modi Script focus on recognizing individual handwritten characters. The datasets used in these studies primarily consisted of individual characters. In contrast, our work is the first to tackle the challenge of \textit{direct transliteration} of handwritten Modi script to Devanagari as shown in Figure \ref{fig:dataset_figures}. 
The dataset curation was particularly challenging due to the scarcity of accessible Modi script documents and the limited number of experts capable of accurately transliterating these documents into Devanagari. In addition, we have created a synthetic dataset named SynthMoDe Dataset (see Appendix \ref{appendix:synthmode}). A contrast between these two datasets highlights the importance of the proposed MoDeTrans dataset, which handles the real-world challenges of Modi script transliteration.
 
2. Our second major contribution is the vision-language model (VLM) based MoScNet framework for transliterating  Modi script images into Devanagari script. MoScNet is inspired by the widespread success of large language models (LLMs) in various downstream tasks. Our key observation is that the transliteration task does not require the computational complexity of full-scale LLMs. Based on this, MoScNet leverages the knowledge distillation (KD) technique, whereby a smaller student model learns transliteration from a ``low-rank adaptation'' (LoRA) adapted pre-trained teacher model. The use of KD allows MoScNet to remain lightweight and computationally efficient, making it suitable for deployment in low-resource environments.

3. Innovative Transformer Architecture: We propose a new decoder-based transformer design within the student model, featuring \textit{parallel attention} and \textit{QK-normalization}. To the best of our knowledge, this is the first application of these techniques in a causal modeling setting, significantly enhancing the model's representational capabilities.

4. Superior Performance over Baselines:  We conducted comprehensive evaluations spanning diverse architectures—from recurrent neural networks (RNNs) to LLMs. 
In this analysis, the \textbf{MoScNet-XL} configuration, incorporating \textbf{429 million parameters} and guided by a \textbf{LoRA}-adapted, pre-trained \textbf{LlaMA-3 70B} teacher model, demonstrated the best overall performance. This result represents a \textit{notable improvement} compared to all previously evaluated architectures. Similarly, on the optical character recognition (OCR) task, our model performs comparably with the recent models and outperforms previous models.

5. Advancement of Script Transliteration Research: By providing an enriched dataset, a high-performance transliteration model, and a novel transformer-based student architecture, our work lays a strong foundation for future research in script transliteration, resource creation, and related natural language processing tasks. Our framework offers significant utility to historians and researchers. Our technique promotes digitizing and preserving fragile historical documents, ensuring their accessibility for future generations.

\section{Background and Related Work}\label{related_work}
\textbf{Challenges of Modi script recognition:} 
Modi script character recognition presents several challenges \cite{theoretical}. The script's cursive nature, diverse writing styles, and issues like angular strokes, broken lines, blurring, and noise make accurate recognition difficult. A key challenge in handwritten optical character recognition (HOCR) is the need for large, well-annotated datasets to train robust models capable of handling these variations. Segmentation is crucial in overcoming these challenges \cite{recentadvance}. Researchers have explored various statistical and machine-learning approaches to address the challenges of HOCR. Building on this, \cite{10775029} identifies ten specific constraints that further complicate Modi script recognition, such as character similarities and non-uniformity. These factors underscore the difficulty of developing a reliable HOCR system for ancient manuscripts.

\textbf{Relevant datasets:} Over the years, several datasets have been developed to support the digitization and recognition of the Modi script. The MODI-HHDoc dataset \cite{hhdoc} consists of 3350 documents written in Modi script, which seeks to promote digitization, recognition, transcription, and transliteration systems. Datasets like the MODI-HChar dataset \cite{hchar}, Handwritten Modi Lipi Barakhadi dataset \cite{barakhadi}, and Handwritten Modi Characters dataset \cite{hwmodichars}  classify individual characters. While these character-focused datasets have improved recognition accuracy, they do not address the challenge of transliterating complete Modi script texts.  
Kundaikar et al. \cite{kundaikar2024modi} use a synthetic dataset for Modi script transliteration. However, a synthetic dataset may not capture historical complexities like writing styles and vocabulary variations. 

\textbf{Related works:} Researchers have explored various techniques for Modi script recognition, focusing primarily on character identification and word transcription. The approaches steadily advanced from traditional methods to modern deep-learning techniques. The early approaches used Zernike moments and zoning techniques \cite{zernike}. The rotational invariance of Zernike moments made them more effective than Hu's seven invariant moments, leading to improved accuracy in character recognition. Similarly, \cite{otsu} introduced a combination of Otsu's binarization algorithm and the Kohenen neural network for Modi script recognition. They use zoning to improve the efficiency of the template matching technique. 
 
As deep learning gained traction, \cite{cnnsvm} proposed using an autoencoder to extract the features for Modi character recognition and SVM for Modi script character classification. 
A few works have used CNN architectures or machine learning models for Modi Script character recognition and word transcription \cite{transcription,alexnet,xception,sayyedpreserving}. While these models demonstrated promising results, they were primarily limited to recognizing individual characters rather than complete words or sentences.

Kundaikar et al. \cite{kundaikar2024modi} used an LSTM designed using OCRopus, for transliterating the synthetically generated Modi script images into Devanagari text. Sawant et al. \cite{transcription} demonstrated strong performance in individual word transcription. 
 However, these approaches face challenges when applied to real-world Modi script documents, which have complex handwriting styles with multiple lines of cursive, angular text and no space between the words. These characteristics make segmentation and accurate transcription particularly complex.

\section{Proposed MoDeTrans Dataset} \label{sec:MoDeTrans}
\begin{figure}[ht]
    \centering
\includegraphics[width=0.45\textwidth]{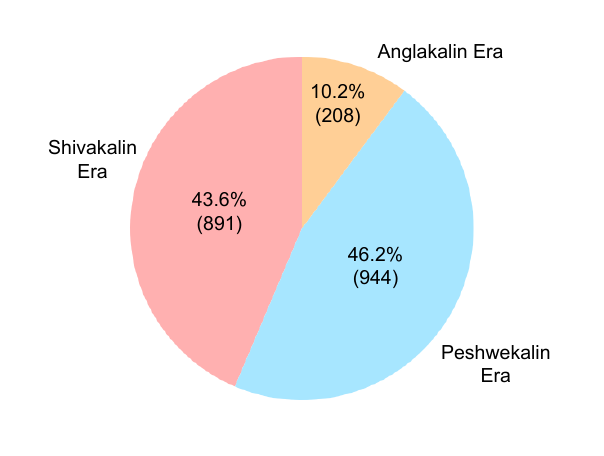}
    \caption{Era-wise distribution of images in the MoDeTrans dataset  }     
    \label{fig:modi_stats}
\end{figure}

\begin{figure}[htbp]
    \centering    \includegraphics[width=0.98\textwidth]{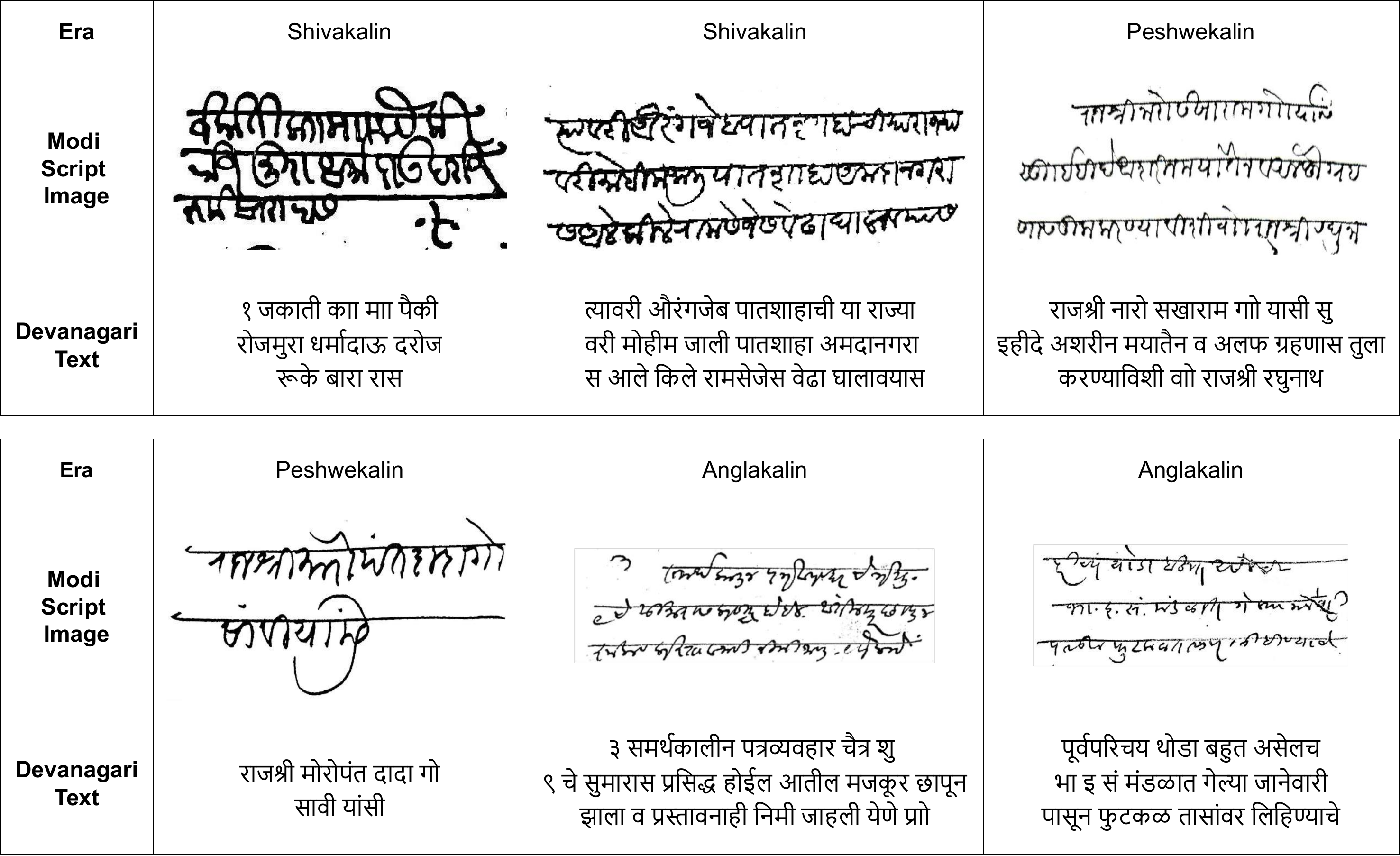}  
    \caption{Modi script images from different eras with Devanagari transliteration from MoDeTrans Dataset
    }
    \label{fig:dataset_figures}
\end{figure}

We introduce the MoDeTrans (\textbf{Mo}di script to \textbf{De}vanagari \textbf{Trans}literation) dataset, containing 2043 images of Modi script and their corresponding Devanagari text transliterations. The dataset encompasses three significant eras: the Shivakalin era, the Peshwekalin era, and the Anglakalin era. In the Shivakalin Era, Modi script fonts were very diverse and much more rudimentary than the more sophisticated and standardized fonts of the Peshwekalin Era. Figure~\ref{fig:modi_stats} shows the era-wise distribution of the number of images. MoDeTrans includes different writing styles of Modi script, such as \textit{Chitnisi}, \textit{Mahadevpanti}, \textit{Bilavalkari}, and \textit{Ranadi} \cite{pandey2011modi}. Figure~\ref{fig:dataset_figures} illustrates the MoDeTrans dataset, showcasing sample images and their corresponding text transliterations categorized by era.

The main reason for including Modi script images only from the Shivakalin, Peshwekalin, and Anglakalin eras is that well-preserved documents from the Adyakalin and Yadavkalin eras are scarce. Many manuscripts from these earlier periods have been lost or damaged over time. Also, the Modi script has changed over time, making it challenging to include all variations without expert knowledge or extra resources. Some old documents are hard to read due to faded ink or worn-out pages, making accurate transliteration challenging. 

\textbf{Preprocessing: }
Initially, each image was segmented into 3–4 parts, each containing approximately 3–4 lines and an average word count of 40. The images were then converted to grayscale, followed by a noise removal process. For most images, we utilized bilateral filtering with a kernel size of 3 to effectively remove noise while preserving edge details. However, certain images, particularly those containing Modi script from old historical records, presented unique challenges due to their age and condition. To address these, we manually handled the noise using Photoshop tools, allowing for a more precise restoration process according to the specific characteristics of these images. Once noise removal was complete, Gaussian adaptive thresholding was applied to enhance text visibility (see Appendix \ref{appendix:preprocessing}). Finally, a deskewing process was carried out to correct any misalignment in the images, ensuring optimal readability for further processing.

\begin{figure}[ht]
    \centering    
     \includegraphics[width=0.9\textwidth]{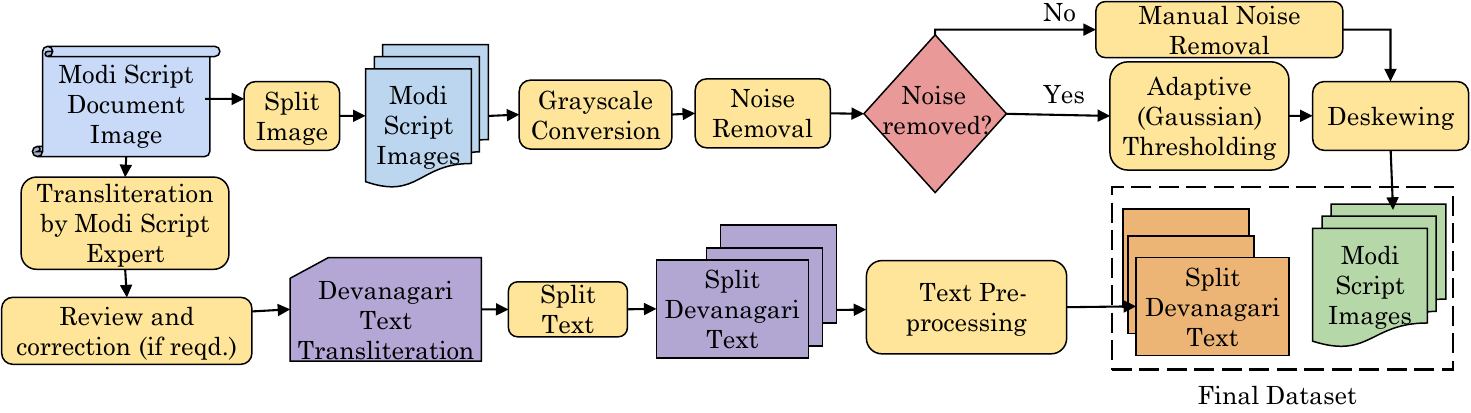}
    \caption{Steps involved in the creation of the MoDeTrans dataset}
    \label{fig:dataset_flowchart}
\end{figure}

For the text preprocessing flow, the text files were first split according to their corresponding Modi script images. Special characters and punctuation were then removed to ensure cleaner and more structured data for further processing. Figure \ref{fig:dataset_flowchart} outlines the preprocessing steps for dataset creation.

 \textbf{Challenges in dataset collection:} The data collection task was difficult as the Modi script documents are available only with institutes like the archives department, the Institute of Oriental Studies, the Bharat Itihas Sanshodhan Mandal in Pune, and a few historians. 
 We have taken 924 images from the Modi-HHDOC dataset \cite{hhdoc}. Additionally, 216 images were sourced from archives, and Modi script experts generously contributed 903 images.
It is worth highlighting that all the images in the Modi-HHDOC dataset only depict Modi script documents. The corresponding transliterations for these documents were meticulously generated manually by our Modi script experts to ensure accuracy and authenticity. 
 Manual transliteration posed significant challenges due to the limited number of experts capable of accurately transliterating.

\textbf{Preserving authenticity: }
The dataset comprises images from credible institutions like Maharashtra's history and archives departments.
The dataset includes only genuine historical data (not some randomly/synthetically written text), thus complying with the ethical standards. Hence, this dataset provides a reliable bridge for exploring Maharashtra's rich historical legacy.
We adopted several measures to ensure our dataset's highest integrity and quality. Expert transliterators with significant experience in Modi Script were engaged to accurately transliterate historical documents, minimizing any risk of distortion that could lead to misinterpretation. Clear guidelines were provided to ensure an objective and faithful representation of the original text and to eliminate potential biases. Additionally, Modi Script experts conducted a cross-verification process to identify and rectify any errors in manual transliteration. These measures collectively ensured that the dataset remains authentic and reliable, preserving Maharashtra's historical records with high fidelity.

\textbf{Train-test-split:} The dataset was divided into training, testing, and validation sets in an 80:10:10 ratio. The images from each era were proportionally split across these sets to ensure fair distribution.
We took explicit and rigorous measures to prevent overlap between the train-test-validation sets. The dataset was manually reviewed and verified by a team of 10 annotators, who ensured no shared samples across the training, testing, and validation splits. This manual intervention provided an additional layer of scrutiny beyond automated splitting mechanisms, further mitigating the risk of unintended data leakage.

\section{Proposed MoScNet Framework}\label{sec:methodology}

\textbf{Key idea:} We propose a novel framework, MoScNet (\textbf{Mo}di \textbf{Sc}ript \textbf{Net}work) based on the concept of Knowledge Distillation \cite{knowledgeDistiallation} to transliterate the Modi script images into Devanagari text. This approach is lightweight for inference and computationally efficient, as a relatively small student model is trained to capture rich knowledge from a larger teacher model. Additionally, the LoRA \cite{lora}  technique is employed to fine-tune the teacher model, further enhancing its performance and, consequently, that of the student model. In our framework, the teacher is a pre-trained LLM, such as LlaMA 2 \cite{llama2} or Mistral 7B \cite{mistral}, whereas the student model is an LLM with fewer parameters.

\subsection{Proposed Architecture}
Given a Modi script image as the input, we aim to transliterate it into Devanagari text. This requires extracting meaningful features from the image and converting those features into corresponding text. Vision-Language Models (VLMs) are well-suited for such image-to-text tasks. However, directly fine-tuning pre-trained VLMs such as LLaVA \cite{llava} or Pixtral \cite{agrawal2024pixtral} for transliteration from Modi script images to Devanagari text is challenging because these models are primarily trained on English-language datasets.

We propose a KD-based architecture, where a smaller model is trained under the supervision of a larger, pretrained language model (LLM). This approach employs a teacher-student paradigm, where the teacher model guides and supervises the student model in transliterating Modi Script images into Devanagari text. In our proposed architecture, the teacher shares response-based knowledge with the student, meaning the teacher model's final predictions guide and supervise the student Model. The rationale for adopting KD is that the transliteration task does not necessitate the computational complexity of large LLMs, such as variants of LlaMA \cite{llama} or Mistral. Instead, lightweight and computationally efficient models can effectively handle this task. Our experimental results provide evidence supporting this approach. 

\begin{figure*}[htbp]
    \centering  \includegraphics[width=0.95\textwidth]{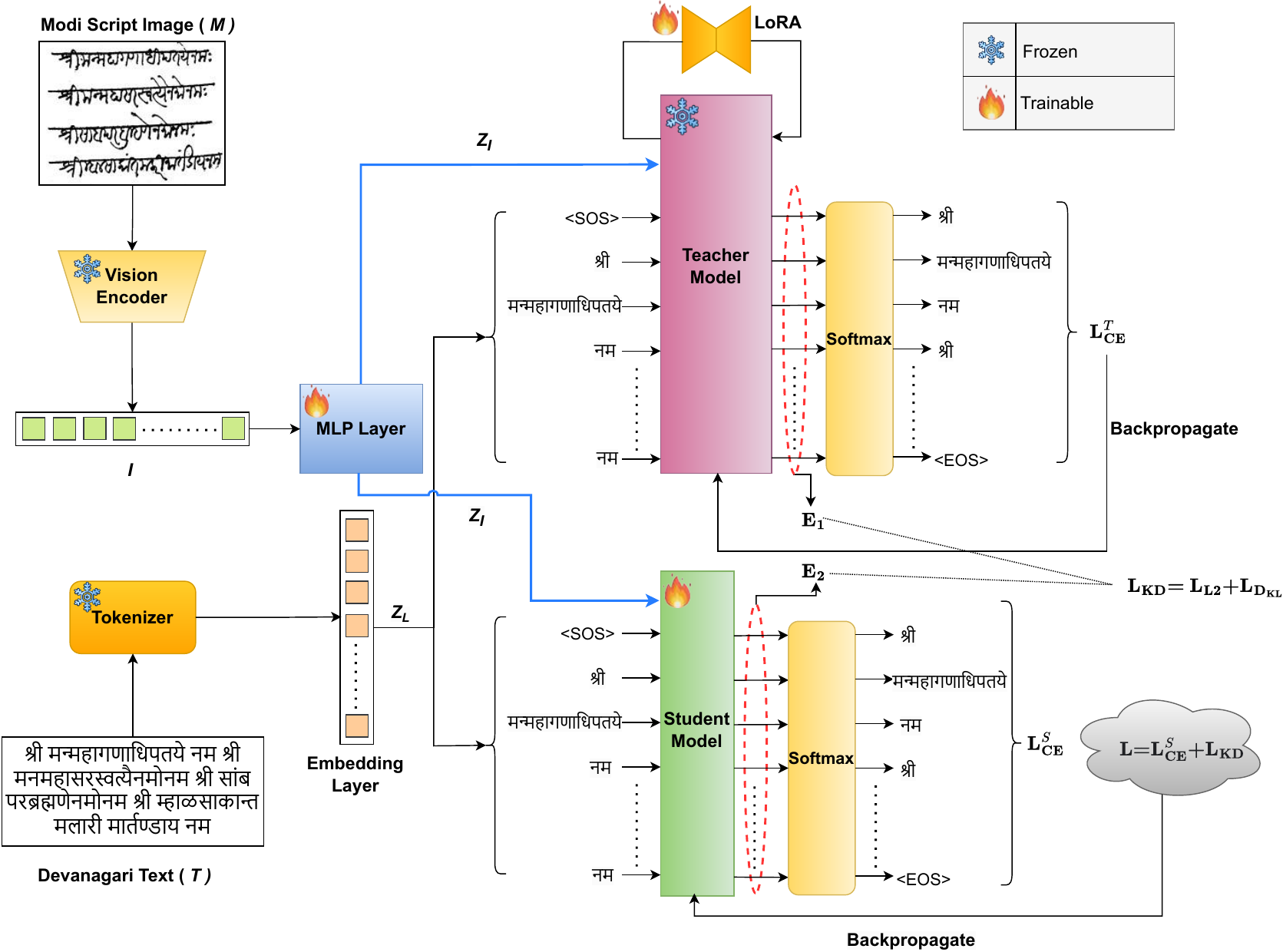}  
    \caption{MoScNet  architecture diagram showing teacher-student knowledge distillation Model}
    \label{fig:architecture_diagram}
\end{figure*}

Figure \ref{fig:architecture_diagram} shows our proposed MoScNet architecture. We now explain its key components. 

\textbf{Vision Encoder:} A vision encoder processes and converts input Modi script image $M$ into image representations $I$. These image representations are then passed to a trainable MLP layer. The MLP layer transforms $I$ into language embeddings $Z_I$.
\begin{align}
I &= \text{VisionEncoder}(M) \notag \\
Z_I &= \text{MLP}(I)
\label{eq:vision_encoder}
\end{align}

\textbf{Language Encoder: }The Devanagari text transliteration $T$ is first processed by a pre-trained tokenizer, which splits the sequence into tokens.
We employed the same Byte-Pair Encoding (BPE) tokenizer used by our teacher model without introducing any additional preprocessing steps for the Devanagari script. Our rationale is grounded in the observation that several Large Language Models (LLMs) such as LLaMA-2, LLaMA-3, and Phi-3 Mini already incorporate training on Hindi data—an Indian language that shares the same Devanagari script as Marathi. Because Marathi and Hindi share the same orthographic system, the existing vocabulary and tokenization scheme capture Marathi text significantly. Consequently, we decided not to expand the vocabulary or retrain the tokenizer specifically for Marathi, thus avoiding unnecessary complexity and preserving the benefits of a well-established subword vocabulary learned from large-scale pretraining corpora.
These tokens are then passed through an embedding layer, generating the corresponding language embeddings $Z_L$. The image embeddings $Z_I$, obtained from Equation~\ref{eq:vision_encoder}, are concatenated with the language embeddings $Z_L$ to form combined embeddings $Z_C$. These combined embeddings $Z_C$ are subsequently used as input to both the teacher and student models.
\begin{align}
Z_C &= [ Z_I \odot Z_L ]
\label{eq:llm_input}
\end{align}

\textbf{Teacher Model: }In our approach, the teacher model is a pretrained LLM, such as Mistral 7B, LlaMA 2, or LlaMA 3 \cite{llama3}, fine-tuned for the task of transliterating Modi Script images into Devanagari text. This fine-tuning enhances the teacher model's ability to improve the performance of the student model. Instead of directly fine-tuning the teacher model by updating its weights, we adopt an adapter-based approach like LoRA (Low-Rank Adaptation). The LoRA technique is employed to fine-tune pre-trained models by freezing their weight matrices and introducing trainable low-rank decomposition matrices at each layer of the transformer-based architecture. This matrix decomposition enhances the model's performance by significantly reducing the number of trainable parameters. These trainable low-rank matrices serve as modular LoRA components, which can be easily swapped for different tasks, ensuring zero additional inference latency.

We freeze the teacher model's weight matrices and train only the low-rank LoRA matrices on the dataset. This significantly reduces the number of trainable parameters while maintaining performance.
The teacher model takes \( Z_C \) as input (Eq.~\ref{eq:llm_input}) and generates hidden representations \( E_1 \). Subsequently, a softmax function is applied to produce a probability distribution over the next possible tokens. The cross-entropy loss, \(  L_{CE}^T \), is backpropagated to guide the trainable LoRA module adjust its weights effectively. Let $N$ be the total number of tokens in the sequence, $y_t$ be the true token at position $t$ and $\hat{y}_t$ be the predicted probability for the true token $y_t$ at position $t$.
\begin{align}
{L_{CE}^T} = - \frac{1}{N} \sum_{t=1}^{T} \log \hat{y}_t
\end{align}

\begin{figure}[ht]
\centering
\includegraphics[width=0.45\textwidth]{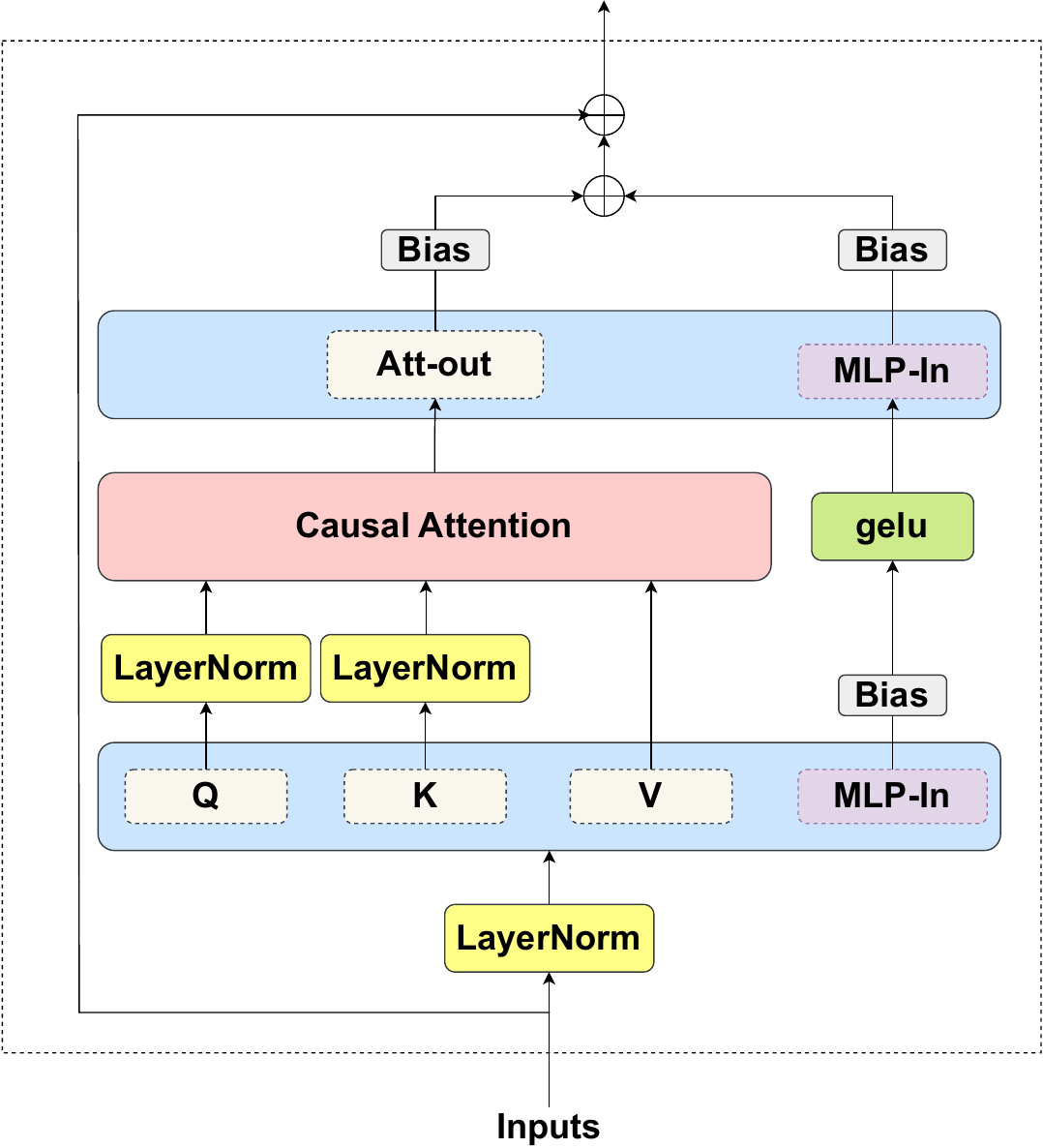}  
    \caption{Architecture of student model - A decoder layer with parallel Attention-MLP blocks}
    \label{fig:student_model}
\end{figure}

\textbf{Student Model: } The student model architecture is inspired from \cite{scalingvit22}. Figure~\ref{fig:student_model} shows a single decoder block of the student model; the full model stacks $N$ such blocks. Unlike the full attention mechanism proposed in \cite{scalingvit22}, the student model incorporates \textit{causal attention} to suit its task better.
The student model processes input representations $Z_C$ and produces hidden representations $E_2$. The learning process involves knowledge distillation, where the student minimizes the difference between its representations ($E_2$) and those of the teacher model ($E_1$). The overall loss $L$ is a weighted combination of \textit{cross-entropy loss} ($L_{CE}^S$) that governs the primary learning task, \textit{L2 norm loss}($L_{L2}$) that measures the Euclidean distance between $E_1$ and $E_2$ and \textit{KL-divergence loss}($L_{D_{KL}}$) that captures the discrepancy between the probabilistic distributions of $E_1$ and $E_2$. Including KL-divergence enables the student to better mimic the distribution of the teacher's internal representations, thereby improving the quality of knowledge transfer.
\begin{equation}
L = L_{CE}^S + L_{KD} = L_{CE}^S + L_{L2} + L_{D_{KL}}
\label{eq:final_student_loss}
\end{equation}

\section{Experimental Results}
 
\textbf{Experimental platform:} To train MoScNet, we employed an initial learning rate of 1e-6 for the teacher models and 1e-4 for the student models. We utilized the AdamW optimizer for training both teacher and student, combined with a cosine learning rate scheduler to dynamically adjust the learning rate during training. We set the batch size to 4 per GPU, with gradient accumulation steps of 8 to effectively increase the batch size and stabilize training. For the teacher LLM, we adapted LoRA with a rank of 64 to enhance efficiency. Input images were resized to dimensions of 128 x 256 using a vision encoder prior to encoding. The entire training was performed on 16 x 8 Nvidia H100 GPUs, leveraging the high computational efficiency of BFloat16 (BF16) precision. We have also calculated $CO_2$ emmission rate in section \ref{appendix:co2emm}

\subsection{MoScNet results on transliteration task}
\textbf{Quantitative results:} We present a comprehensive quantitative comparison of various transliteration methods on the proposed MoDeTrans and SynthMoDe Dataset (see Appendix \ref{appendix:synthmode}) in Table~\ref{tab:quantitative_analysis}. The evaluation includes Connectionist Temporal Classification (CTC)-based architectures such as CRNN~\cite{crnn} and VITSTR~\cite{vitstr}, as well as a range of LLMs, including multiple configurations of LlaMA-2 and LlaMA-3, and the Phi3 Mini model with SigLip \cite{siglip} as the Vision Encoder. To assess transliteration quality, we employ the BLEU score~\cite{bleu} (see Appendix~\ref{appendix:evaluation_metric}).

 \begin{table}[htbp]
\centering
\caption{BLEU scores ($\uparrow$) for different methods on proposed MoDeTrans Dataset and Synthetic SynthMoDe Dataset.}
\begin{tabular}{|C{2cm}|c|C{2.1cm}|C{2cm}|C{3cm}|}
\hline
 Model & Vision Encoder & MoDeTrans & SynthMoDe & Zero-shot evaluation on SynthMoDe \\
\hline
CRNN & - & 27.33   & 28.65  & 27.92\\ 
VITSTR & - & 31.21    & 33.42  & 32.17\\
LlaMA-2 7B & SigLIP   & 36.54   & 39.78  & 37.98\\
LlaMA-2 13B & SigLIP      & 40.22   & 42.11 & 41.56\\
Phi3 Mini & SigLIP      & 39.97    & 41.48 & 40.28\\
LlaMA-3 8B & SigLIP   & 43.11   & 46.17 & 44.19 \\
LlaMA-3 70B & SigLIP    & 49.37  &52.52  & 49.98\\
\hline
\multicolumn{5}{|c|}{\textbf{With Proposed Method}} \\
\hline
Student Model & Teacher Model & MoDeTrans  & SynthMoDe &  Zero-shot Evaluation on SynthMoDe  \\
\hline
MoScNet-XL & LlaMA-2 7B   & 38.19   & 40.98  & 40.02\\
MoScNet-XL & LlaMA-2 13B   & 41.17  & 42.92 & 41.29\\
MoScNet-XL & Phi3 Mini   & 40.76    & 43.38 & 41.45\\
MoScNet-XL & LlaMA-3 8B   & 45.12   & 48.75 & 46.62\\
MoScNet-XL & LlaMA-3 70B   & \textbf{51.03}   &\textbf{53.97} & \textbf{51.51} \\
\hline
\end{tabular}

\label{tab:quantitative_analysis}
\end{table}

Conventional CTC-based approaches, such as CRNN and VITSTR, exhibit limited capacity to capture the nuanced visual-linguistic patterns inherent in the transliteration task. Their lower BLEU scores reflect limited effectiveness in mapping character sequences from the source image to the target text. In contrast, LLMs such as LlaMA-3 70B leverage extensive pre-training to acquire rich, high-level representations of complex patterns. To enable more effective learning under limited computational and data constraints, we employ LoRA adapters for Parameter-Efficient Fine Tuning (PEFT). As a result, these models substantially outperform earlier methods, with LlaMA-3 70B achieving a BLEU score of 49.37 and 52.52 on MoDeTrans and SynthMode datasets, respectively.

Our proposed \textbf{MoScNet-XL} framework, guided by a knowledge-distilled LlaMA-3 70B teacher model, further advances the state-of-the-art. Incorporating a novel student model architecture that employs \emph{parallel attention} mechanisms, MoScNet-XL efficiently integrates the distilled knowledge from the teacher model while using only 429M parameters. This parallel attention design enables the student model to better align visual and textual features, improving pattern recognition and a more robust understanding of the transliteration process.

The results substantiate the advantages of KD combined with an innovative transformer-based architecture for the student model. MoScNet-XL achieves a BLEU score of 51.03 and 53.97 on MoDeTrans and SynthMode Dataset, respectively, surpassing all previously reported approaches and establishing a new benchmark for transliteration accuracy. This improvement highlights the potential of integrating large-scale teacher models with carefully designed student architectures to enhance performance on visually grounded language tasks.

The results further demonstrate that the BLEU score on the SynthMoDe dataset is higher than that on the MoDeTrans dataset. This difference can be attributed to the relatively simple nature of the synthetic SynthMoDe dataset compared to the more complex, real-world MoDeTrans dataset.  
Notably, the zero-shot evaluation—where the model was trained on the MoDeTrans dataset and tested on the SynthMoDe dataset—yielded competitive results compared to training the model explicitly on the SynthMoDe dataset. This observation highlights the richness of the MoDeTrans dataset, which encompasses complex writing styles, diverse fonts, and patterns from different eras. These characteristics contributed to the strong zero-shot performance on the SynthMoDe dataset.

\begin{figure*}[htbp]
    \centering
        \caption{Qualitative analysis of transliteration using different methods: correctly transliterated text is shown in black, while errors are highlighted in red}
    \includegraphics[width=\textwidth]{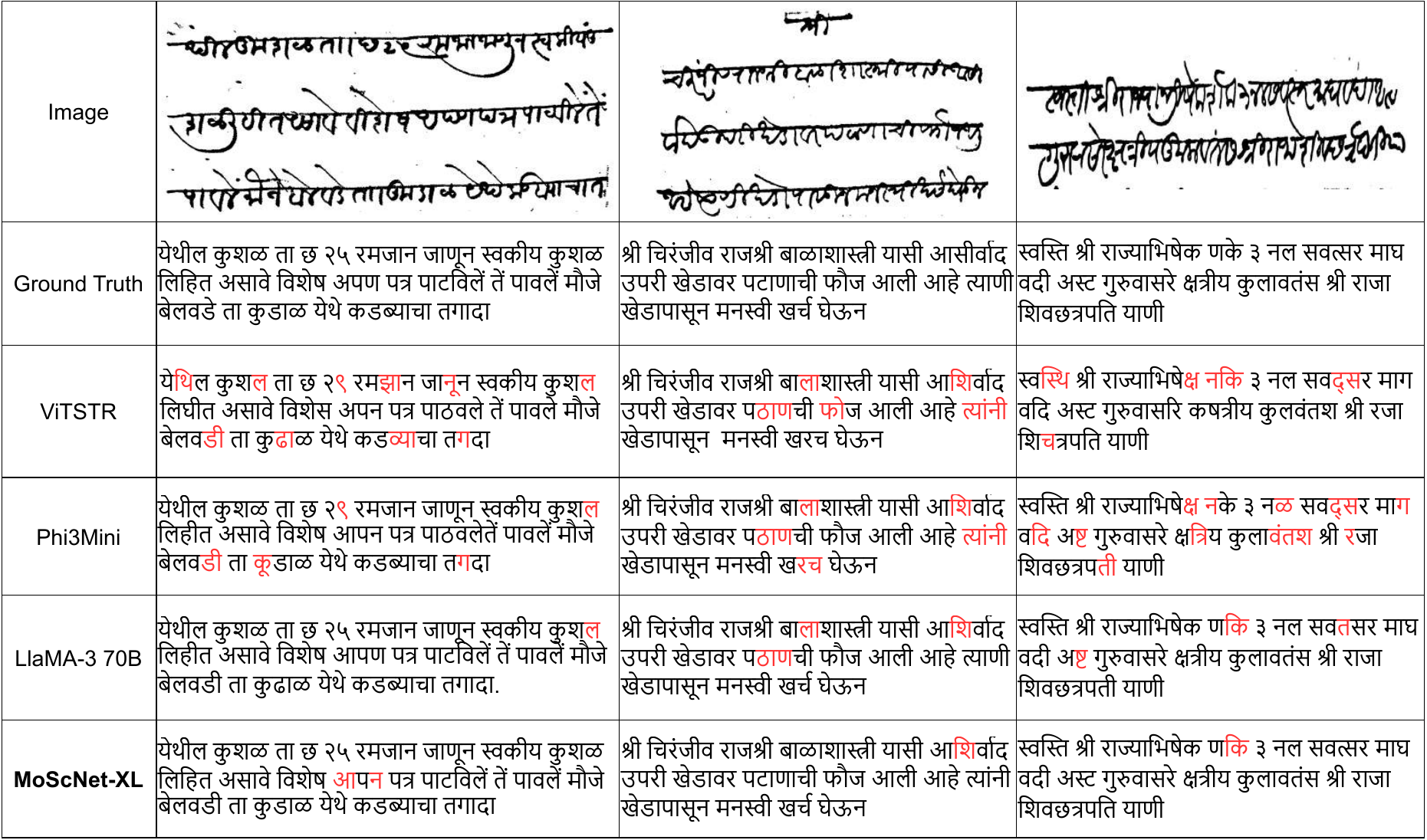}    \label{fig:qualitative_analysis}
\end{figure*}

\textbf{Qualitative results:} Figure~\ref{fig:qualitative_analysis} showcases the transliteration performance of various methods. Incorrectly transliterated characters are highlighted to illustrate deviations from the ground truth, providing insight into each approach's error patterns and relative strengths. Notice that Phi3Mini and LlaMA-3 70B provide better results than VITSTR. This discrepancy can be attributed to the superior ability of large language models (LLMs) to capture complex patterns and features in images, which VITSTR lacks. While LlaMA-3 70B and MoScNet-XL demonstrate competitive performance, our proposed framework, MoScNet-XL, outperforms all other methods, underscoring the effectiveness of knowledge distillation in enhancing transliteration accuracy.

While the MoScNet-XL model demonstrates strong overall performance in transliteration tasks, it exhibits minor inconsistencies. Notably, it struggles to accurately differentiate between visually similar characters such as {\dn B} (bha) and {\dn m} (ma), {\dn k} (ka) and {\dn P} (pha), as well as {\dn V} (ṭa), {\dn Y} (ṭha), and {\dn W} (ḍha). Furthermore, the model occasionally fails to distinguish between {\dn n} (na) and {\dn Z} (ṇa) and encounters difficulties with vowel diacritics, such as {\dn k\?} (ke) and {\dn Ek} (ki). Additionally, in some instances, the model omits the anusvāra or the dot above certain characters, resulting in incorrect phonetic representations. Addressing these discrepancies through further refinement would significantly enhance the model’s transliteration accuracy and reliability.

\subsection{MoScNet results on scene text OCR task} 
\textbf{Quantitative results:} Our proposed model, MoScNet, achieves state-of-the-art performance not only in ModiScript transliteration but also in Scene Text Optical Character Recognition (OCR) tasks. To demonstrate MoScNet's versatility, we evaluated its performance on widely recognized OCR benchmarks, ICDAR13 \cite{karatzas2013icdar} and ICDAR15 \cite{karatzas2015icdar}, after training on the MJSynth dataset \cite{Jaderberg14}.

\begin{table}[ht]
\centering
\caption{Character-level accuracy on ICDAR13 and ICDAR15 datasets}
\label{tab:ocr_icdar_comparison}
\begin{tabular}{|l|c|c|}
\hline
\textbf{Methods} & \textbf{ICDAR13} & \textbf{ICDAR15} \\
\hline
CRNN & 90.90 & 70.80 \\
ABINet & 95.00 & 79.00 \\
ParSeq-A & 96.20 & 82.20 \\
CoFormer-A & 96.32 & 82.91 \\
MoScNet-XL & \textbf{97.89} & \textbf{85.52} \\
\hline
\end{tabular}
\end{table}

As shown in Table \ref{tab:ocr_icdar_comparison}, MoScNet-XL significantly surpasses recent OCR models like CRNN, ABINet \cite{fang2021read}, ParSeq-A \cite{bautista2022scene}, and CoFormer-A \cite{Deshmukh_2024_WACV}. On ICDAR13, MoScNet-XL attains the highest character level accuracy of \textbf{97.89\%}, outperforming CoFormer-A by 1.57 percentage point. Similarly, on the more challenging ICDAR15 dataset, characterized by complex backgrounds and irregular text patterns, MoScNet-XL achieves character level accuracy of \textbf{85.52\%}, an improvement of 2.61 percentage point over CoFormer-A.

MoScNet's superior performance is attributed to its multi-scale convolutional architecture, effectively capturing local and global contextual features, combined with advanced attention mechanisms and robust sequence modeling capabilities. These results underscore MoScNet's effectiveness as a versatile OCR model for both transliteration and traditional scene-text recognition tasks.

\begin{figure*}[ht]
    \centering
        \caption{Qualitative Results for Scene Text OCR}
    \includegraphics[width=\textwidth]{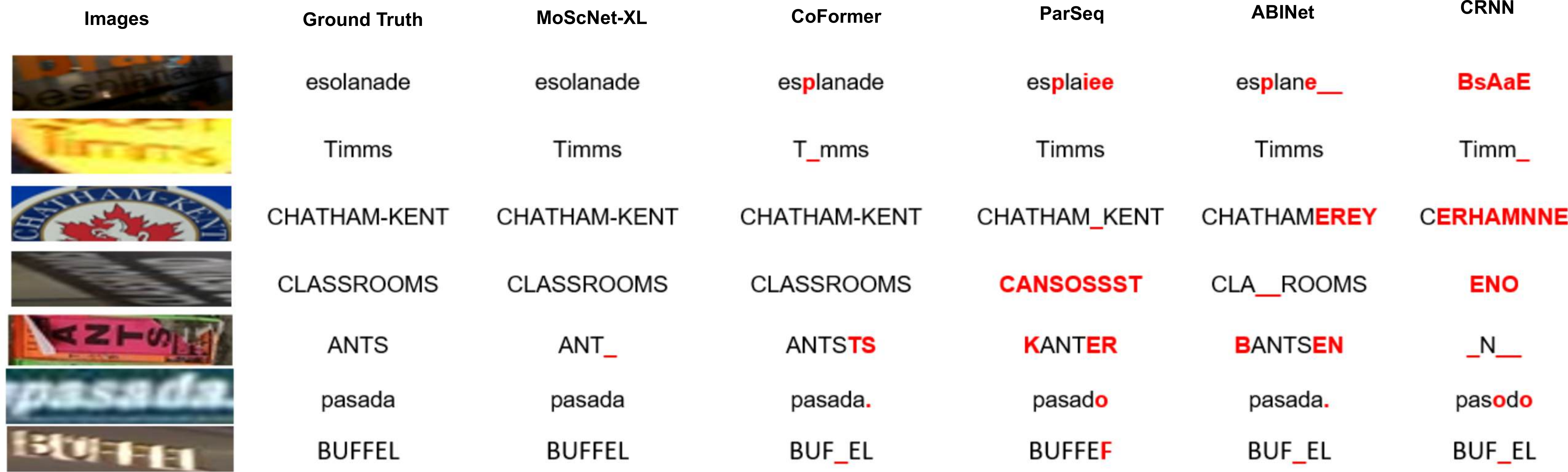}    \label{fig:ocr_qualitative_analysis}
\end{figure*}

\textbf{Qualitative results:} Figure \ref{fig:ocr_qualitative_analysis} presents the qualitative results. 
These examples demonstrate MoScNet-XL's superior recognition accuracy across various challenging real-world text instances. These results highlight MoScNet-XL’s robustness and superior generalization ability for complex scene text recognition tasks. The model effectively handles variations in background clutter, distortions, and irregular text alignments, making it a highly reliable OCR solution.

MoScNet-XL accurately predicts words like esolanade, CHATHAM-KENT, CLASSROOMS, pasada and BUFFEL, whereas other models make errors such as character substitutions or missing letters, deletions, or extraneous punctuation. In fact, CRNN completely misinterprets the text. For ANTS, MoScNet-XL has a minor mistake (ANT\_), yet other methods generate additional incorrect characters or  recognize the word completely incorrectly (e.g., KANTER in ParSeq, BANTSEN in ABINet).

\section{Ablation Studies}

\subsection{Ablation based on vision encoder}
We evaluated MoScNet's performance by substituting different vision encoders, namely CLIP \cite{clip}, ViT-L \cite{vit}, SwinTransformer \cite{swintransformer}, and SigLIP, while holding MoScNet-XL as the student model (see Section~\ref{ablation:size_of_student_model}) and LlaMA-3 70B as the teacher model. As reported in Table~\ref{tab:vision_encoder_ablation}, SigLIP consistently outperformed other vision encoders. This superior performance can be attributed to SigLIP's ability to represent image features as discrete, semantically meaningful tokens. By effectively bridging continuous image representations and discrete textual embeddings, SigLIP facilitates more robust cross-modal alignment and concept-level understanding. As a result, the model can more easily integrate visual cues with language-based reasoning, leading to enhanced multi-modal learning outcomes.
\begin{table}[htbp]
\centering
\caption{Ablation results with different vision encoders}
\begin{tabular}{|l|c|}
\hline
\textbf{Vision Encoder} & \textbf{BLEU Score} \\
\hline
CLIP              & 49.32 \\
ViT-L             & 50.11 \\
SwinTransformer    & 50.67 \\
SigLIP            & 51.03 \\
\hline
\end{tabular}

\label{tab:vision_encoder_ablation}
\end{table}

\subsection{Ablation Based on Size of Student Model}\label{ablation:size_of_student_model}
We evaluate the performance of our framework, MoScNet by varying the size of the Student Model. As illustrated in Figure~\ref{fig:student_model}, the student Model is constructed by stacking \(N\) identical blocks. The ablation study focuses on the impact of Hidden Size, Number of Blocks, and Parameter Count on the model's performance.
Table~\ref{tab:MoScNet_comparison} highlights that MoScNet-XL, with 429M parameters, outperforms all other variants in this study. We utilized SigLIP as the vision encoder and LlaMA-3 70B as the teacher model for this evaluation. The results reaffirm the efficacy of Knowledge Distillation in producing a compact and lightweight MoScNet-XL model. Remarkably, this smaller model performs better than the LlaMA-3 70B model despite being 163 times smaller in parameter count.
\begin{table*}[htbp]
\centering
\caption{Comparison of different MoScNet variants on BLEU score and throughput (T/s = tokens per second).}
\begin{tabular}{|l|c|c|c|c|c|}
\hline
\textbf{Method} & \textbf{Hidden Size} & \textbf{\#Blocks} & \textbf{Param.} & \textbf{BLEU Score} & \textbf{Throughput (T/s)} \\
\hline
MoScNet-S  & 256  & 6  & \textbf{44M}   & 43.32 & \textbf{100} \\
MoScNet-M  & 512  & 13 & 92M   & 48.36 & 84  \\
MoScNet-L  & 768  & 18 & 234M  & 50.04 & 67  \\
MoScNet-XL & 1024 & 21 & 429M  & \textbf{51.03} & 44  \\
\hline
\end{tabular}

\label{tab:MoScNet_comparison}
\end{table*}

\subsection{Ablation based on the  loss function}  
We experimented with various loss functions as shown in Table~\ref{tab:loss_ablation}. 
We observe that the student model trained exclusively with $L_{CE}^S$ loss achieves a BLEU score of 48.02, which is lower than the full MoScNet-XL (51.03). This finding demonstrates that while the $L_{CE}^S$-only approach benefits from the quality of the MoDeTrans dataset, it does not reach the performance of our proposed method.

Also, incorporating both $L_{L2}$ and $L_{D_{KL}}$ losses significantly enhances the student model's performance compared to using either loss individually.  
\begin{table}[htbp]
\centering
\caption{Ablation results with different loss functions }
\begin{tabular}{|l|c|}
\hline
\textbf{Loss Function} & \textbf{BLEU Score} \\
\hline
$L_{CE}^S$       & 48.02    \\
$L_{CE}^S$ + $L_{L2}$        & 50.41    \\
$L_{CE}^S$ + $L_{D_{KL}}$         & 50.29 \\
\textbf{$\textbf{L}_\textbf{CE}^\textbf{S}$}+\textbf{$\textbf{L}_{\textbf{L2}} + \textbf{L}_{\textbf{D}_{\textbf{KL}}}$ }     & \textbf{51.03} \\
\hline
\end{tabular}
\label{tab:loss_ablation}
\end{table}

\subsection{Ablation Based on Student Architecture}

As shown in Table~\ref{tab:student_model_ablation}, the removal of parallel attention within the MoScNet-XL student model—a key aspect of our transformer-based approach—reduces the BLEU score from its best performance, indicating that parallel attention layers more effectively capture and integrate long-range dependencies. Similarly, omitting the QK-Norm decreases the BLEU score, suggesting that normalizing the attention keys and queries helps stabilize the attention mechanism, improving feature alignment and representation consistency. Together, these findings underline that both parallel attention and QK-Norm are integral to achieving optimal transliteration accuracy in our framework.

\begin{table}[htbp]
\centering
\caption{Ablation results based on the student architecture.}
\begin{tabular}{|l|c|}
\hline
\textbf{Student architecture} & \textbf{BLEU Score} \\ \hline
W/O Parallel Attention                    & 49.22          \\
W/O QK-Norm                               & 50.07          \\ \hline
\end{tabular}
\label{tab:student_model_ablation}
\end{table}

\section{Conclusion}
We introduced MoDeTrans, the first dataset providing Modi script images and their corresponding transliteration in Devanagari. We proposed the Modi script transliteration framework using KD, which converts Modi script text in document images to the Devanagari script. Our work is the first to perform direct transliteration from the Handwritten Modi script to the Devanagari script. Our proposed MoScNet outperforms previous methods on this task.

\textbf{Acknowledgement}

We sincerely thank Mrs. Kanchan Karai, a distinguished expert in Modi script, for her invaluable support in this research. Her expertise in transliterating historical Modi script documents into Devanagari script was instrumental in creating a comprehensive dataset for our study. Her contributions have greatly enriched our work, and we deeply appreciate her assistance.
\bibliographystyle{splncs04}
\bibliography{bibliography}
\appendix
\section{Appendix}
\label{sec:appendix}

\subsection{SynthMoDe Dataset}\label{appendix:synthmode}
\begin{figure*}[ht]
    \centering
        \caption{SynthMoDe Dataset Images: Image 1 showcases the MarathiCursive font, while Image 2 features the Noto-Sans-Modi font with corresponding Devanagari transliteration.
}
    \includegraphics[width=\textwidth]{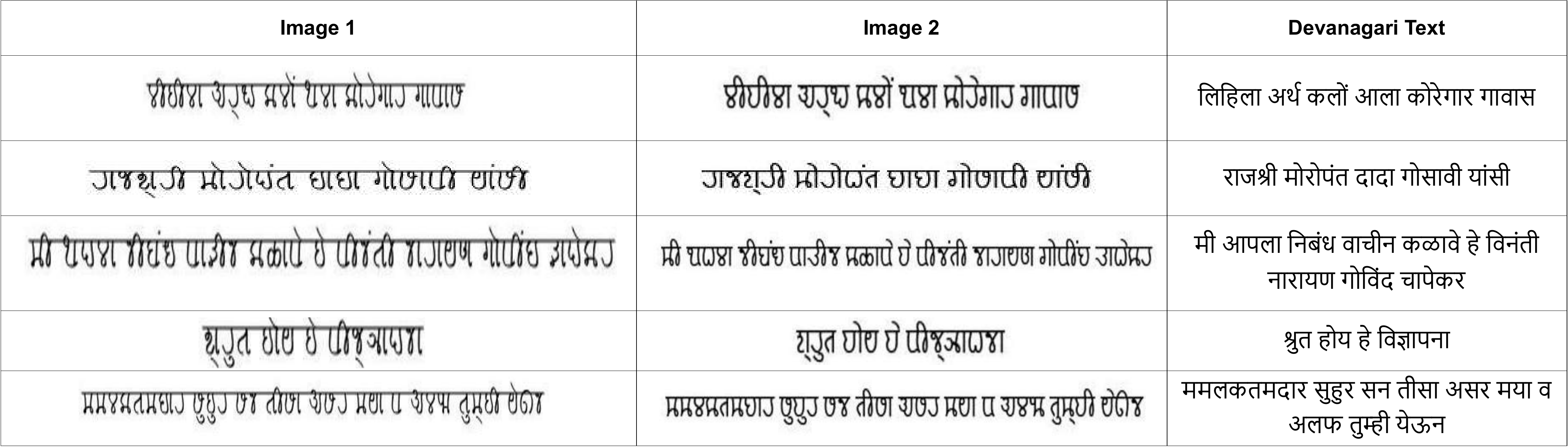}    \label{fig:synthetic_dataset}
\end{figure*}

We have also created a synthetic dataset, named SynthMoDe (\textbf{Synth}etic \textbf{Mo}di Script to \textbf{De}vanagari) by following the steps mentioned in \cite{kundaikar2024modi}. The need for creation of Synthetic Dataset arises because the authors of \cite{kundaikar2024modi} have not made their dataset publicly available. In addition, it was necessary to highlight the challenges and complexities involved in the transliteration of real handwritten Modi Script documents to Devanagari compared to synthetic Modi Script documents. The synthetic Modi Script images are generated from Devanagari Text transliterations of our proposed MoDeTrans Dataset \ref{sec:MoDeTrans}. We have created two synthetic images for each Devanagari text by using  MarathiCursive \cite{marathicursive} and Noto-Sans-Modi \cite{notosansmodi} fonts, respectively. The SynthMoDe Dataset is available at \url{https://huggingface.co/datasets/historyHulk/SynthMoDe}. Figure~\ref{fig:synthetic_dataset} depicts sample images with their transliterations from the SynthMoDe Dataset.

\subsection{Gaussian Thresholding}\label{appendix:preprocessing}

The Gaussian method comes under the Adaptive type of thresholding. Here, Gaussian blur is initially applied to the image before calculating the threshold value. This operation is as follows:

\begin{equation}
    \text{t} = \text{k} \times \sigma
\end{equation}
Where:
\begin{align*}
    t & : \text{threshold value}. \\
    k & : \text{constant factor } (0.5 < k < 2). \\
    \sigma & : \text{standard deviation of Gaussian blur}.
\end{align*}

After obtaining the threshold value \textit{t},  pixels with intensity greater than the threshold \textit{t} are set to white (255 pixel value), whereas pixels with intensity values lower than or equal to threshold \textit{t} are set to black (0 pixel value).The type of thresholding to be used was decided based on the quality of the image after applying both methods individually. 
 
\subsection{Evaluation Metric}\label{appendix:evaluation_metric}
BLEU(Bilingual Evaluation Understudy) score \cite{bleu} is used as an evaluation metric for reviewing the model's performance. The formula of the BLEU score is given by:
\begin{equation}
    \text{BLEU-score} = \text{bp} \times \exp\left(\sum_{i=1}^{N} w_i \cdot \log p_i\right)
    \label{eq:bleu}
\end{equation}

Where:\\
\begin{align*}
    & \text{BLEU}: \text{BLEU score}. \\
    & \text{bp}: \text{Brevity Penalty}. \\
    & w_i: \text{Weights for n-gram precisions}. \\
    & p_i: \text{n-gram precision}.
\end{align*}

The formula for n-gram precision (\( p_i \)) based on \cite{bleu} is given by:

\begin{equation}
    p_i = 
    \frac{
        \begin{aligned}
            &\text{Count of n-grams that match} \\
            &\text{in the candidate and references}
        \end{aligned}
    }{
        \text{Total n-gram count of candidate}
    }
\end{equation}

Where candidate is machine-generated text and references are the desired output text. 
The Brevity Penalty (bp) formula \cite{bleu} is given by:

\begin{equation}
\text{bp} = 
\begin{cases}
1 & \text{if } len_c > len_r \\
e^{(1 - len_r/len_c)} & \text{if } len_c \leq len_r
\end{cases}
\end{equation}

Where:
\begin{align*}
    len_c & : \text{Length of the candidate text}. \\
    len_r & : \text{Length of the closest reference text}.
\end{align*}

\subsection{Estimated \texorpdfstring{$CO_2$}{CO2} Emissions Calculation}\label{appendix:co2emm}
\noindent To estimate the $CO_2$ emissions produced by running 16 nodes of 8 Nvidia H100 GPUs (totaling 128 GPUs), we proceed as follows:

\begin{enumerate}
    \item \textbf{Assumptions:}
    \begin{itemize}
        \item Power consumption per Nvidia H100 GPU: $\approx 700\,\text{W} = 0.7\,\text{kW}$
        \item Number of GPUs: $128$
        \item Carbon intensity of electricity: $\approx 0.45\,\text{kgCO}_2/\text{kWh}$ (regional average)
    \end{itemize}

    \item \textbf{Total Power Consumption:} For 128 GPUs:
    \[
    P_{\text{total}} = 128 \times 0.7\,\text{kW} = 89.6\,\text{kW}
    \]

    \item \textbf{Energy Use Over One Hour:}
    \[
    E = P_{\text{total}} \times t = 89.6\,\text{kW} \times 1\,\text{h} = 89.6\,\text{kWh}
    \]

    \item \textbf{CO$_2$ Emissions:}
    Given a carbon intensity $C = 0.45\,\text{kgCO}_2/\text{kWh}$:
    
    \[
    \text{Emissions per hour} = E \times C\] 

    \[
    \text{Total Emission} = 89.6\,\text{kWh} \times 0.45\,\frac{\text{kgCO}_2}{\text{kWh}} 
    \]

    \[\approx 40.32\,\text{kg CO}_2/\text{hour}\]

\end{enumerate}

\noindent Hence, operating 128 Nvidia H100 GPUs for 1.3 hours of training may produce approximately $52\,\text{kg}$ of CO$_2$ emissions, depending on the actual power draw and the regional energy mix.

\section{Limitations and Ethics Statement}
The scope of this work can be further enhanced by expanding our existing datasets to include additional data. A larger dataset would significantly improve the BLEU score, giving more accurate results. Currently, the dataset comprises data from three eras: Shivakalin, Peshwekalin, and the Anglakalin era. However, there is potential to include data from the Adyakalin and Yadavkalin eras, subject to the availability of relevant documents.

Our work strictly complies with ethical guidelines.  
No data containing sensitive or personal information was used in this work. For the transliteration of the documents, the Modi script experts were selected solely based on their qualifications and expertise in this domain. No human bias was involved. The transliterators guaranteed that the documents were correctly transliterated to the Devanagari script, and there was no loss of information in this process.

\end{document}